\documentclass[12pt]{article}
\usepackage{fullpage}
\usepackage{pdfsync}
\usepackage{amssymb,latexsym,amsthm}
\usepackage{amsmath, url, algorithm, algorithmic, verbatim}
\newtheorem{theorem}{Theorem}
\newtheorem{proposition}{Proposition}
\newtheorem{definition}{Definition}
\newtheorem{lemma}{Lemma}

\newcommand{\Comments}{1}
\newcommand{\note}[2]{\ifnum\Comments=1\textcolor{#1}{#2}\fi}

\def\mechname{\textnormal{CLM}}
\def\S{\mathcal{S}}
\def\s{\mathbf{s}}

\def\profit{\mathtt{Profit}}
\renewcommand\P{\mathcal{P}}
\newcommand{\kl}{\textnormal{KL}}
\newcommand{\relint}{\textnormal{relint}}
\def\worstcaseloss{\text{WorstCaseLoss}}

\def\x{\mathbf{x}}

\def\p{\mathbf{p}}
\def\q{\mathbf{q}}
\def\r{\mathbf{r}}
\def\v{\mathbf{v}}
\def\1{\mathbf{1}}
\def\0{\mathbf{0}}

\def\e{\mathbf{e}}

\def\E{\mathbb{E}}

\def\reals{\mathbb{R}}

\def\H{\mathcal{H}}

\def\w{\mathbf{w}}
\def\dom{\text{dom}}
\def\y{\mathbf{y}}

\def\O{\mathcal{O}}

\def\cost{\mathtt{Cost}}
\def\payout{\mathtt{Payout}}
\def\rhob{\boldsymbol{\rho}}

\def\pbar{\boldsymbol{\bar p}}

\newcommand{\argmin}{\mathop{\rm argmin}}
\newcommand{\argmax}{\mathop{\rm argmax}}

\title{A Collaborative Mechanism for Crowdsourcing Prediction Problems}

\author{
Jacob Abernethy \\%\thanks{ Use footnote for providing further information
Division of Computer Science \\
University of California at Berkeley\\
\texttt{jake@cs.berkeley.edu} \\
\and
Rafael M. Frongillo \\
Division of Computer Science \\
University of California at Berkeley\\
\texttt{raf@cs.berkeley.edu} \\ 
}

\begin{document}
	
\maketitle  

\begin{abstract}
	Machine Learning competitions such as the Netflix Prize have proven reasonably successful as a method of ``crowdsourcing'' prediction tasks. But these competitions have a number of weaknesses, particularly in the incentive structure they create for the participants. We propose a new approach, called a Crowdsourced Learning Mechanism, in which participants collaboratively ``learn'' a hypothesis for a given prediction task. The approach draws heavily from the concept of a prediction market, where traders bet on the likelihood of a future event. In our framework, the mechanism continues to publish the current hypothesis, and participants can modify this hypothesis by wagering on an update. The critical incentive property is that a participant will profit an amount that scales according to how much her update improves performance on a released test set.
\end{abstract}

\section{Introduction}

The last several years has revealed a new trend in Machine Learning: prediction and learning problems rolled into prize-driven competitions. One of the first, and certainly the most well-known, was the \emph{Netflix prize} released in the Fall of 2006. Netflix, aiming to improve the algorithm used to predict users' preferences on its database of films, released a dataset of 100M ratings to the public and asked competing teams to submit a list of predictions on a test set withheld from the public. Netflix offered \$1,000,000 to the first team achieving prediction accuracy exceeding a given threshold, a goal that was eventually met. This competitive model for solving a prediction task has been used for a range of similar competitions since, and there is even a new company (\url{kaggle.com}) that creates and hosts such competitions. Such prediction competitions have proven quite valuable for a couple of important reasons: (a) they leverage the abilities and knowledge of the public at large, commonly known as ``crowdsourcing'', and (b) they provide an incentivized mechanism for an individual or team to apply their own knowledge and techniques which could be particularly beneficial to the problem at hand. This type of prediction competition provides a nice tool for companies and institutions that need help with a given prediction task yet can not afford to hire an expert. The potential leverage can be quite high: the Netflix prize winners apparently spent more than \$1,000,000 in effort on their algorithm alone.

Despite the extent of its popularity, is the Netflix competition model the ideal way to ``crowdsource'' a learning problem? We note several weaknesses:

\paragraph{It is anti-collaborative.} Competitors are strongly incentivized to keep their techniques private. This is in stark contrast to many other projects that rely on crowdsourcing -- Wikipedia being a prime example, where participants must build off the work of others. Indeed, in the case of the Netflix prize, not only do leading participants lack incentives to share, but the work of non-winning competitors is effectively wasted.

\paragraph{The incentives are skewed and misaligned.} The winner-take-all prize structure means that second place is as good as having not competed at all. This ultimately leads to an equilibrium where only a few teams are actually competing, and where potential new teams  never form since catching up seems so unlikely. In addition, the fixed achievement benchmark, set by Netflix as a 10\% improvement in prediction RMSE over a baseline, leads to misaligned incentives. Effectively, the prize structure implies that an improvement of \%9.9 percent is worth nothing to Netflix, whereas a 20\% improvement is still only worth  \$1,000,000 to Netflix. This is clearly not optimal.

\paragraph{The nature of the competition precludes the use of proprietary methods.} By requiring that the winner reveal the winning algorithm, potential competitors utilizing non-open software or proprietary techniques will be unwilling to compete. By participating in the competition, a user must effectively give away his intellectual property.

In this paper we describe a new and very general mechanism to crowdsource prediction/learning problems.  Our mechanism requires participants to place bets, yet the space they are betting over is the set of \emph{hypotheses} for the learning task at hand. At any given time the mechanism publishes the current hypothesis $\w$ and participants can wager on a modification of $\w$ to $\w'$, upon which the modified $\w'$ is posted. Eventually the wagering period finishes, a set of test data is revealed, and each participant receives a payout according to their bets. The critical property is that every trader's profit scales according to how well their modification \emph{improved the solution} on the test data.

The framework we propose has many qualities similar to that of an \emph{information} or \emph{prediction market}, and many of the ideas derive from recent research on the design of \emph{automated market makers} \cite{H03,H07,CP07,CV10,ACV11}. Many information markets already exist; at sites like \url{Intrade.com} and \url{Betfair.com}, individuals can bet on everything ranging from election outcomes to geopolitical events. There has been a burst of interest in such markets in recent years, not least of which is due to their potential for combining large amounts of information from a range of sources. In the words of Hanson et al \cite{hanson2006information}: ``Rational expectations theory predicts that, in equilibrium, asset prices will reflect all of the information held by market participants. This theorized information aggregation property of prices has lead economists to become increasingly interested in using securities markets to predict future events.'' In practice, prediction markets have proven impressively accurate as a forecasting tool~\cite{LHI09,Ber:01,Wol:04}.

The central contribution of the present paper is to take the framework of a prediction market as a tool for information aggregation and to apply this tool for the purpose of ``aggregating'' a hypothesis (classifier, predictor, etc.) for a given learning problem. The crowd of ML researchers, practitioners, and domain experts represents a highly diverse range of expertise and algorithmic tools. In contrast to the Netflix prize, which pitted teams of participants against each other, the mechanism we propose allows for everyone to contribute whatever knowledge they may have available towards the final solution. In a sense, this approach decentralizes the process of solving the task, as individual experts can potentially apply their expertise to a subset of the problem on which they have an advantage.
Whereas a market price can be thought of as representing a consensus estimate of the value of an asset, our goal is to construct a consensus hypothesis reflecting all the knowledge and capabilities about a particular learning problem\footnote{It is worth noting that Barbu and Lay utilized concepts from prediction markets to design algorithms for classifier aggregation \cite{florida}, although their approach was unrelated to crowdsourcing.}.

\paragraph{Layout:} We begin in Section~\ref{sec:gsr} by introducing the simple notion of a \emph{generalized scoring rule} $L(\cdot,\cdot)$ representing the ``loss function'' of the learning task at hand. In Section~\ref{sec:clm} we describe our proposed Crowdsourced Learning Mechanism (CLM) in detail, and discuss how to structure a CLM for a particular scoring function $L$, in order that the traders are given incentives to minimize $L$. In Section~\ref{sec:compression} we give an example based on the design of Huffman codes. In Section~\ref{sec:prediction} we discuss previous work on the design of prediction markets using an \emph{automated prediction market maker} (APMM).
We observe that any APMM is just a particular CLM and, moreover, we fully classify what types of problems can be solved with an APMM.
In Section~\ref{sec:examples} we finish by considering two learning settings (e.g. linear regression) and we construct a CLM for each. The proofs have been omitted throughout, but these are available in the full version of the present paper.

\paragraph{Notation:} Given a smooth strictly convex function $R : \reals^d \to \reals$, and points $\x,\y \in \dom(R)$, we define the \emph{Bregman divergence} $D_R(\x,\y)$ as the quantity $R(\x) - R(\y) - \nabla R(\y)\cdot(\x - \y)$. For any convex function $R$, we let $R^*$ denote the \emph{convex conjugate} of $R$, that is $R^*(\y) := \sup_{\x \in \dom(R)} \y \cdot \x - R(\x)$. We shall use $\Delta(S)$ to refer to the set of integrable probability distributions over the set $S$, and $\Delta_n$ to refer to the set of probability vectors $\p \in \reals^n$. The function $H : \Delta_n \to \reals$ shall denote the \emph{entropy function}, that is $H(\p) := - \sum_{i=1}^n \p(i) \log \p(i)$.  We use the notation $\kl(\p;\q)$ to describe the \emph{relative entropy} or \emph{Kullback-Leibler divergence} between distributions $\p,\q \in \Delta_n$, that is $\kl(\p;\q) := \sum_{i=1}^n \p(i) \log \frac{\p(i)}{\q(i)}$. We will also use $\e_i \in \reals^n$ to denote the $i$th standard basis vector, having a 1 in the $i$th coordinate and 0's elsewhere.

% 
% We give a full overview of the details of our mechanism below, but we pause now to discuss several properties of the proposed framework that are highly beneficial yet not present in the winner-take-all Netflix approach.
% 
% \begin{itemize}
% 	\item \textbf{The mechanism is collaborative.} By ``trading'' in this competition, participants must reveal, in a certain sense, their internal knowledge about the problem. The posted hypothesis $\w$ can be thought of as a ``consensus'' estimate of the optimal, and not only are participants given this current estimate, they are in fact only rewarded for improving the solution relative to $\w$.
% 	\item \textbf{The mechanism matches up the incentives of the creator and the participants.} By setting the payout structure to tie the financial rewards to the participant's relative improvement in the loss (according to the outcome $X$), the goals of the creator and the incentives of the participants are matched up perfectly.
% 	\item \textbf{The mechanism incentivizes partial improvement.} Whereas the winner-take-all model provides a payout to only the winning team, our mechanism provides incentives for any improvement. That is, if a particular participant has some private knowledge about the problem, say the irrelevance of certain attributes, then it can still earn money for acting on this information without having a winning algorithm
% \end{itemize}

\section{Scoring Rules and Crowdsourced Learning Mechanisms} \label{sec:gsr_clm}

We shall now provide a full description of our proposed crowdsourced learning mechanism. We begin by discussing the notion of a \emph{scoring rule}, a well-studied object from statistics for the purpose eliciting ``good'' probability forecasts \cite{gneiting2007strictly}. We propose a weaker notion which we call a \emph{generalized scoring rule} $L(\cdot,\cdot)$ which shall reflect the loss function of the learning problem at hand. We then proceed to describe the CLM framework, and in particular we present the important case when a CLM \emph{implements} a generalized scoring rule $L$. We provide a range of properties and results for $L$-CLMs. 

\subsection{Generalized Scoring Rules} \label{sec:gsr}

For the remainder of this section, we shall let $\H$ denote some set of \emph{hypotheses}, which we will assume is a convex subset of $\reals^n$. We let $\O$ be some arbitrary set of \emph{outcomes}. We use the symbol $X$ to refer to either an element of $\O$, or a random variable taking values in $\O$.

We recall the notion of a \emph{scoring rule}, a concept that arises frequently in economics and statistics~\cite{gneiting2007strictly}.
\begin{definition}
  Let $\P \subseteq \Delta(\O)$ be some convex set of distributions on an outcome space $\O$.  A \emph{scoring rule} is a function $S:\P\times\O\to\reals$ where, for all $P\in\P$, $P \in \argmax_{Q\in\P} \E_{X\sim P} S(Q,X)$.
\end{definition}\vspace{-4pt}
In other words, if you are paid $S(P,X)$ upon stating belief $P\in\P$ and outcome $X$ occurring, then you maximize your expected utility by stating your true belief. We offer a much weaker notion:
\begin{definition}
  Given a convex hypothesis space $\H \subset \reals^n$ and an outcome space $\O$, let $L:\H\times\O\to\reals$ be a continuous function. Given any $P \in \Delta(\O)$, let \[W_L(P) := \argmin_{\w\in\H} \E_{X \sim P} [L(\w; X)].\] Then we say that $L$ is a \emph{Generalized Scoring Rule} (GSR) if $W_L(P)$ is a nonempty convex set for every $P \in \Delta(\O)$.
\end{definition}\vspace{-4pt}

The generalized scoring rule shall represent the ``loss function'' for the learning problem at hand, and in Section~\ref{sec:clm} we will see how $L$ is utilized in the mechanism. The hypothesis $\w$ shall represent the advice we receive from the crowd, $X$ shall represent the test data to be revealed at the close of the mechanism, and $L(\w;X)$ shall represent the loss of the advised $\w$ on the data $X$. Notice that we do not define $L$ to be convex in its first argument as this does not hold for many important cases. Instead, we require the weaker condition that $\E_X [L(\w; X)]$ is minimized on a convex set for any distribution on $X$.

Our scoring rule differs from traditional scoring rules in an important way.  Instead of starting with the desire know about the true value of $X$, and then designing a scoring rule which incentivizes participants to elicit their belief $P \in \P$, our objective is precisely to minimize our scoring rule.  In other words, traditional scoring rules were a means to an end (eliciting $P$) but our generalized scoring rule is the end itself.  One can recover the traditional scoring rule definition by setting $\H=\P$ and imposing the constraint that $P\in W_L(P)$.

A useful class of GSRs $L$ are those based on a Bregman divergence.
\begin{definition}
  \label{def:div-based}
  We say that a GSR $L:\H\times\O\to\reals$ is \emph{divergence-based} if there exists an alternative hypothesis space $\H' \subset \reals^m$, for some $m$, where we can write
  \begin{equation}
    L(\w;X) \equiv D_R(\rho(X), \psi(\w)) + f(X)
    \label{eq:div-based}
  \end{equation}
  for arbitrary maps $\rho: \O \to \H', f :\O \to \reals$, and $\psi: \H\to\H'$, and any closed strictly convex $R: \H' \to \reals$ whose convex conjugate $R^*$ is finite on all of $\reals^m$.
\end{definition}\vspace{-4pt}

This property allows us to think of $L(\w;X)$ as a kind of distance between $\rho(X)$ and $\psi(\w)$.  Clearly then, the minimum value of $L$ for a given $X$ will be attained when $\psi(\w) = \rho(X)$, given that $D_R(\x,\x) = 0$ for any Bregman divergence.  In fact, as the following proposition shows, we can even think of the expected value $\E[L(\w;X)]$, as a distance between $\E[\rho(X)]$ and $\psi(\w)$.

\begin{proposition}
  Given a divergence-based GSR $L(\w;X) = D_R(\rho(X), \psi(\w)) + f(X)$ and a belief distribution $P$ on $\O$, we have $W_L(P) = \psi^{-1}\bigl(\E_{X \sim P}\, [\rho(X)]\bigr)$.
  \label{prop:mean-min}
\end{proposition}\vspace{-4pt}
\begin{proof}
  All expectations in the following are over $X\sim P$.  Expanding $L$, we have
  \begin{align*}
    W_L(P)
    &= \argmin_{\w\in\H} \Bigl\{ \E\, \bigl[R(\rho(X)) - R(\psi(\w))
    - \nabla R(\psi(\w)) \cdot (\rho(X) - \psi(\w)) + f(X)\bigr] \Bigr\}
    \\
    &= \argmin_{\w\in\H} \Bigl\{ \E\, [R(\rho(X)) + f(X)] - R(\psi(\w))
      - \nabla R(\psi(\w)) \cdot (\E[\rho(X)] - \psi(\w)) \Bigr\}
    \\
    &= \argmin_{\w\in\H} \Bigl\{ R(\E[\rho(X)]) - R(\psi(\w))
      - \nabla R(\psi(\w)) \cdot (\E[\rho(X)] - \psi(\w)) \Bigr\}
    \\
    &= \argmin_{\w\in\H} \Bigl\{ D_R(\E[\rho(X)],\psi(\w)) \Bigr\}
    \hspace{4pt} = \hspace{4pt}
    \psi^{-1}\bigl(\E[\rho(X)]\bigr)
  \end{align*}
  where the last line follows from the strict convexity of $R$ and properties of divergences.
\end{proof}

We now can see that the divergence-based property greatly simplifies the task of minimizing $L$; instead of worrying about $\E[L(\cdot; X)]$ one can simply base the hypothesis directly on the expectation $\E[\rho(X)]$.  As we will see in section~\ref{sec:prediction}, this also leads to efficient prediction markets and crowdsourcing mechanisms.
% 
% Note though that $\rho$ could map to a higher-dimensional space, thus
% potentially revealing more information about the bidder's belief $P$.
% For example, $\rho:X\mapsto(X,X^2,X^3,\dots,X^k)$ would incentivize
% bidders to reveal the first $k$ moments of their distribution.  In
% fact, if we had $\rho:\O\to\P$ such that $\E_P[\rho(X)] = P$, we would
% recover the scoring rules of Gneiting and
% Rafferty~\cite{gneiting2007strictly}.

%  The proof is straightforward and can
% be found in Banerjee et al. \cite{banerjee2005clustering} as well as
% other sources.

\subsection{The Crowdsourced Learning Mechanism} \label{sec:clm}

We will now define our actual mechanism rigorously.

\begin{definition}
  A \emph{Crowdsourced Learning Mechanism (\mechname)} is the
  procedure in Algorithm~\ref{alg:clm} as defined by the tuple $(\H,
  \O, \cost, \payout)$. The function $\cost : \H \times \H \to \reals$
  sets the cost charged to a participant that makes a modification to
  the posted hypothesis. The function $\payout : \H \times \H \times
  \O \to \reals$ determines the amount paid to each participant when
  the outcome is revealed to be $X$.
  \label{def:clm}
\end{definition}\vspace{-4pt}

\begin{algorithm}[!ht]
  \begin{algorithmic}[1]
    \STATE Mechanism sets initial hypothesis to some $\w_0 \in \H$
    \FOR{rounds $t = 0, 1, 2, \ldots$}
    \STATE Mechanism posts current hypothesis $\w_t \in \H$ 
    \STATE Some participant places a bid on the update $\w_t \mapsto \w'$
    \STATE Mechanism charges participant $\cost(\w_t,\w')$
    \STATE Mechanisms updates hypothesis $\w_{t+1} \leftarrow \w'$
    \ENDFOR
    \STATE Market closes after $T$ rounds and the outcome (test data) $X \in \O$ is revealed
    \FOR{\textbf{each} $t$}
    \STATE Participant responsible for the update $\w_t \mapsto \w_{t+1}$
    receives $\payout(\w_t,\w_{t+1};X)$
    \ENDFOR
  \end{algorithmic}
  \caption{Crowdsourced Learning Mechanism for $(\H, \O, \cost, \payout)$} \label{alg:clm}
\end{algorithm}

The above procedure describes the process by which participants can provide advice to the mechanism to select a good $\w$, and the profit they earn by doing so.  Of course, this profit will precisely determine the incentives of our mechanism, and hence a key question is: how can we design $\cost$ and $\payout$ so that participants are incentivized to provide good hypotheses? The answer is that we shall structure the incentives around a GSR $L(\w;X)$ chosen by the mechanism designer.
\begin{definition}
  For a \mechname\ $A = (\H,\O,\cost,\payout)$, denote the ex-post profit for the bid $(\w\mapsto\w')$ when the outcome is $X \in \O$ by $\profit(\w,\w';X) := \payout(\w,\w';X) - \cost(\w,\w')$.
  We say that $A$ \emph{implements} a GSR $L:\H'\times\O\to\reals$ if there exists a surjective map $\varphi : \H \to \H'$ such that for all $\w_1, \w_2 \in \H$ and $X \in \O$,
  \begin{equation}
    \profit(\w_1,\w_2; X) = L(\varphi(\w_1); X) - L(\varphi(\w_2); X).
    \label{eq:profit-loss}
  \end{equation}
  If additionally $\H'=\H$ and $\varphi = \text{id}_\H$, we call $A$ an $L$-CLM and say that $A$ is \emph{$L$-incentivized}.
\end{definition}\vspace{-4pt}

When a CLM implements a given $L$, the incentives are structured in order that the participants will work to minimize $L(\w;X)$. Of course, the input $X$ is unknown to the participants, yet we can assume that the mechanism has provided a public ``training set'' to use in a learning algorithm. The participants are thus asked not only to propose a ``good'' hypothesis $\w_t$ but to wager on whether the update $\w_{t-1} \mapsto \w_t$ improves generalization error. It is worth making clear that knowledge of the true distribution on $X$ provides a straightforward optimal strategy.
  % Indeed, any participant with full knowledge of the distribution of $X$  Of course, the  The $L$-incentivized property enables participants to profit from revealing their private belief regarding the choice of a ``good'' hypothesis with respect to the loss $L$.  This claim is justified in the following proposition.
\begin{proposition}
  Given a GSR $L:\H\times\O\to\reals$ and an $L$-CLM $(\cost,\payout)$, any participant who knows the true distribution $P\in\P$ over $X$ will maximize expected profit by modifying the hypothesis to any $\w \in W_L(P)$.
  \label{prop:clm-optimum}
\end{proposition}\vspace{-4pt}
\begin{proof}
  By equation \eqref{eq:profit-loss} we can directly compute the expected profit; for any current hypothesis $\w\in\H$, we have
  \begin{align*}
    \argmax_{\w'\in\H} \E_{X\in P}\bigl[\profit(\w,\w';X)\bigr]
    & =
    \argmax_{\w'\in\H} \E_{X\in P}\bigl[ L(\w;X)-L(\w';X) \bigr]
    \\
    & =
    \argmin_{\w'\in\H} \E_{X\in P}\bigl[ L(\w';X) \bigr] \; = \;
    W_L(P),
  \end{align*}
which completes the proof.
\end{proof}

% We now discuss a few properties regarding the design of a CLM. 

\paragraph{Cost of operating a CLM.}
It is clear that the agent operating the mechanism must pay the participants at the close of the competition, and is thus at risk of losing money (in fact, it is possible he may gain).  How much money is lost depends on the bets $(\w_t\mapsto\w_{t+1})$ made by the participants, and of course the final outcome $X$. The agent has a clear interest in knowing precisely the potential cost -- fortunately this cost is easy to compute. The loss to the agent is clearly the total ex-post profit earned by the participants, and by construction this sum telescopes: $\sum_{t=0}^T \profit(\w_t,\w_{t+1};X) = L(\w_0; X) - L(\w_T; X).$ This is a simple yet appealing property of the CLM: the agent pays only as much in reward to the participants as it benefits from the improvement of $\w_T$ over the initial $\w_0$.  It is worth noting that this value could be \emph{negative} when $\w_T$ is actually ``worse'' than $\w_0$; in this case, as we shall see in section~\ref{sec:compression}, the CLM can act as an insurance policy with respect to the mistakes of the participants. A more typical scenario, of course, is where the participants provide an improved hypothesis, in which case the CLM will run at a cost. We can compute the $\worstcaseloss(\text{$L$-\mechname}) := \max_{\w \in \H, X \in \O} \left( L(\w_0; X) - L(\w; X) \right)$.
Given a budget of size $\$B$, the mechanism can always rescale $L$ in order that $\worstcaseloss(\text{$L$-\mechname}) = B$. This requires, of course, that the $\worstcaseloss$ is finite.

\paragraph{Computational efficiency of operating a CLM.}
We shall say that a CLM has the \emph{efficient computation (EC)} property if both $\cost$ and $\payout$ are efficiently computable functions. We shall say a CLM has the \emph{tractable trading (TT)} property if, given a current hypothesis $\w$, a belief $P \in \Delta(\O)$ and a budget $B$, one can efficiently compute an element of the set
    \begin{equation*}
      \argmax_{\w'\in\H} \Bigl\{ \E_{X\sim P}
      \bigl[\profit(\w,\w',X)\bigr] :\: \cost(\w,\w')\leq B \Bigr\}.
    \end{equation*}
    The EC property ensures that the mechanism operator can run the CLM efficiently.  The TT property says that \emph{participants} can compute the optimal hypothesis to bet on given a belief on the outcome and a budget.  This is absolutely essential for the CLM to successfully aggregate the knowledge and expertise of the crowd -- without it, despite their motivation to lower $L(;)$, the participants would not be able to compute the optimal bet.

\paragraph{Suitable collateral requirements.}
We say that a CLM has the \emph{escrow (ES)} property if the $\cost$ and $\payout$ functions are structured in order that, given any wager $(\w \mapsto \w')$, we have that $\payout(\w,\w';X) \geq 0$ for all $X \in \O$. It is clear that, when designing an $L$-CLM for a particular $L$, the $\payout$ function is fully specified once $\cost$ is fixed, since we have the relation $\payout(\w,\w';X) = L(\w;X) - L(\w';X) + \cost(\w,\w')$ for every $\w,\w'\in\H$ and $X \in \O$. A curious reader might ask, why not simply set $\cost(\w,\w') \equiv 0$ and $\payout \equiv \profit$? The problem with this approach is that potentially $\payout(\w,\w';X) < 0$ which implies that the participant who wagered on $(\w \mapsto \w')$ can be indebted to the mechanism and could default on this obligation. Thus the $\cost$ function should be set in order to require every participant to deposit at least enough collateral in escrow to cover any possible losses. 

\paragraph{Subsidizing with a voucher pool.} One practical weakness of a wagering-based mechanism is that individuals may be hesitant to participate when it requires depositing actual money into the system. This can be allayed to a reasonable degree by including a \emph{voucher pool} where each of the first $m$ participants may receive a voucher in the amount of $\$C$. These candidates need not pay to participate, yet have the opportunity to win. Of course, these vouchers must be paid for by the agent running the mechanism, and hence a value of $mC$ is added to the total operational cost.

\section{Warm-up: Compressing an Unfamiliar Data Stream}
\label{sec:compression}

% \subsection{A Primer on First-Order Compression}

Let us now introduce a particular setting motivated by a well-known problem in information theory. Imagine a firm is looking to do \emph{compression} on an unfamiliar channel, and from this channel the firm will receive a stream of $m$ characters from an $n$-sized alphabet which we shall index by $[n]$. The goal is to select a binary encoding of this alpha in such a way that minimizes the total bits required to store the data, as a cost of \$$1$ is required for each bit.

A first-order approach to encode such a stream is to assign a probability distribution $\q \in \Delta_n$ to the alphabet, and to select an encoding of character $i$ with a binary word of length $\log(1/\q(i))$ (we ignore round-off for simplicity).  This can be achieved using Huffman Codes for example, and we refer the reader to Cover and Thomas (\cite{cover1991elements}, Chapter 5) for more details.  Thus, given a distribution $\q$, the firm pays $L(\q;i) = -\log\q(i)$ for each character $i$.  It is easy to see that if the characters are sampled from some ``true'' distribution $\p$, then the expected cost $L(\q;\p) := \E_{i\sim p}\,[L(\q;i)] = \kl(\p;\q) + H(\p)$, which is minimized at $\q = \p$.  Not knowing the true distribution $\p$, the firm is thus interested in finding a $\q$ with a low expected cost $L(\q;\p)$.

An attractive option available to the firm is to \emph{crowdsource} the task of lowering this cost $L(\cdot;\cdot)$ by setting up an $L$-CLM.  It is reasonably likely that outside individuals have private information about the behavior of the channel and, in particular, may be able to provide a better estimate $\q$ of the true distribution of the characters in the channel.  As just discussed, the better the estimate the cheaper the compression.

We set $\H = \Delta_n$ and $\O = [n]$, where a hypothesis $\q$ represents the proposed distribution over the $n$ characters, and $X$ is some character \emph{sampled uniformly} from the stream after it has been observed.  We define $\cost$ and $\payout$ as
\begin{eqnarray*}
	\cost(\q, \q')  :=  \max_{i \in [n]}\; \log(\q(i) / \q'(i)), &\quad&
	\payout(\q, \q'; i) := \log(\q(i) / \q'(i)) + \cost(\q, \q'),
\end{eqnarray*}
which is clearly an $L$-CLM for the loss defined above.  It is worth noting that $L$ is a divergence-based GSR if we take $R(\q) = -H(\q)$, $\rho(i)=\e_i$, $f\equiv 0$, $\psi\equiv\text{id}_{\Delta_n}$, using the convention $0\log 0 = 0$ (in fact, $L$ is the LMSR).  Finally, the firm will initially set $\q_0$ to be its best guess of $\p$, which we will assume to be uniform (but need not be).

We have devised this payout scheme according to the selection of a single character $i$, and it is worth noting that because this character is sampled uniformly at random from the stream (with private randomness), the participants \emph{cannot} know which character will be released. This forces the participants to wager on the empirical distribution $\hat\p$ of the characters from the stream. A reasonable alternative, and one which lowers the payment variance, is to payout according to the $L(\q; \hat{\p})$, which is also equal to the average of $L(\q;i)$ when $i$ is chosen uniformly from the stream.
% NOTE: Some of this content could be returned if desired:
% not to payout according to $L(\q;i)$ for a single $i$, but instead to payout according to the average $i$. This is equivalent to   is to wager on distributions $\q$ which which have , and will thus be forced to express their beliefs as to the distribution $\p$ from which the character is drawn.  While this approach has a high variance in the payout, the firm can lower the variance by averaging the payout over all, or a large subset, of the  paying out $k$ times for $k$ i.i.d. samples of the stream.  Both of these tricks we will see again in Section~\ref{sec:y-mkt}.

The obvious question to ask is: how does this CLM benefit the firm that wants to design the encoding? More precisely, if the firm uses the final estimate $\q_T$ from the mechanism, instead of the initial guess $\q_0$, what is the trade-off between the money paid to participants and the money gained by using the crowdsourced hypothesis?  At first glance, it appears that this trade-off can be arbitrarily bad: the worst case cost of encoding the stream using the final estimate $\q_T$ is $\sup_{i,\q_T} -\log(\q_T(i)) = \infty$. Amazingly, however, by virtue of the aligned incentives, the firm has a very strong control of its total cost (the CLM cost plus the encoding cost).  Suppose the firm scales $L$ by a parameter $\alpha$, to separate the scale of the CLM from the scale of the encoding cost (which we assumed to be \$1 per bit).  Then given any initial estimate $\q_0$ and final estimate $\q_T$, the expected total cost over $\p$ is
\begin{eqnarray*}
	\text{Total expected cost}
    & = &
    \overbrace{H(\p) + \kl(\p; \q_T)}^{\text{Encoding cost of using $\q_T$ given $\p$}}
    \quad + \quad
    \overbrace{\alpha(\kl(\p;\q_0) - \kl(\p;\q_T))}^{\text{Mechanism's cost of getting advice $\q_T$}}\\  
	& = & H(\p) + (1 - \alpha)\kl(\p; \q_T) + \alpha \kl(\p;\q_0)
\end{eqnarray*}

Let us spend a moment to analyze the above expression.  Imagine that the firm set $\alpha=1$.  Then the total cost of the firm would be $H(\p) + \kl(\p; \q_0)$, which is bounded by $\log n$ for $\q_0$ uniform.  Notice that this expression \emph{does not depend on $\q_T$} -- in fact, this cost precisely corresponds to the scenario where the firm had not set up a CLM and instead used the initial estimate $\q_0$ to encode.  In other words, for $\alpha = 1$, the firm is entirely neutral to the quality of the estimate $\q_T$; even if the CLM provided an estimate $\q_T$ which performed worse than $\q_0$, the cost increase due to the bad choice of $\q$ is recouped from payments of the ill-informed participants.

The firm may not want to be neutral to the estimate of the crowd, however, and under the reasonable assumption that the final estimate $\q_T$ will improve upon $\q_0$, the firm should set $0 < \alpha < 1$ (of course, positivity is needed for nonzero payouts).  In this case, the firm will strictly gain by using the CLM when $\kl(\p;\q_T) < \kl(\p;\q_0)$, but still has some insurance policy if the estimate $\q_T$ is poor.

\section{Prediction Markets as a Special Case}
\label{sec:prediction}

Let us briefly review the literature for the type of prediction markets relevant to the present work. In such a prediction market, we imagine a future event to reveal one of $n$ uncertain outcomes. Hanson \cite{H03,H07} proposed a framework in which traders make ``reports'' to the market about their internal belief in the form of a distribution $\p \in \Delta_n$. Each trader would receive a reward (or loss) based on a function of their proposed belief and the belief of the previous trader, and the function suggested by Hanson was the \emph{Logarithmic Market Scoring Rule} (LMSR). It was shown later that the LMSR-based market is equivalent to what is known as a \emph{cost function based automated market makers}, proposed by Chen and Pennock \cite{CP07}. More recently a much broader equivalence was established by Chen and Wortman Vaughan \cite{CV10} between markets based on cost functions and those based on scoring rules. 

The market framework proposed by Chen and Pennock allows traders to buy and sell \emph{Arrow-Debreu} securities (equivalently: shares, contracts), where an Arrow-Debreu security corresponding to outcome $i$ pays out \$1 if and only if $i$ is realized. All shares are bought and sold through an \emph{automated market maker}, which is the entity managing the market and setting prices.  At any time period, traders can purchase \emph{bundles} of contracts $\r \in \reals^n$, where $\r(i)$ represents the number of shares purchased on outcome $i$. The price of a bundle $\r$ is set as $C(\s + \r) - C(\s)$, where $C$ is some differentiable convex cost function and $\s \in \reals^n$ is the ``quantity vector'' representing the total number of outstanding shares. The LMSR cost function is $C(\s) := \frac 1 \eta \log \left( \sum_{i=1}^n \exp(\eta \s(i)) \right)$.

This cost function framework was extended by Abernethy et al. \cite{ACV11} to deal with prohibitively large outcome spaces. When the set of potential outcomes $\O$ is of exponential size or even infinite, the market designer can offer a restricted number of contracts, say $n$ ($\ll |\O|$), rather than offer an Arrow-Debreu contract for each member of $\O$. To determine the payout structure, the market designer chooses a function $\rhob : \O \to \reals^n$, where contract $i$ returns a payout of $\rho_i(X)$ and, thus, a contract bundle $\r$ pays $\rhob(X) \cdot \r$. As with the framework of Chen and Pennock, the contract prices are set according to a cost function $C$, so that a bundle $\r$ has a price of $C(\s + \r) - C(\s)$. The design of the function $C$ is addressed at length in Abernethy et al., to which we refer the reader.

For the remainder of this section we shall discuss the prediction market template of Abernethy et al. as it provides the most general model; we shall refer to such a market as an Automated Prediction Market Maker. We now precisely state the ingredients of this framework.
\begin{definition}\label{def:apmm}
  An \emph{Automated Prediction Market Maker (APMM)} is defined by a tuple $(\S, \O, \rhob, C)$ where $\S$ is the \emph{share space} of the market, which we will assume to be the linear space $\reals^n$; $\O$ is the set of outcomes; $C : \S \to \reals$ is a smooth and convex cost function with $\nabla C(\S) = \relint ( \nabla C(\S))$ (here, we use $\nabla C(\S) := \{ \nabla C(\s) \: | \: \s \in \S \}$ to denote the \emph{derivative space} of $C$); and $\rhob : \O \to \nabla C(\S)$ is a payoff function
  % \footnote{The conditions that $\rhob(\O) \subseteq \nabla C(\S)$ and $\nabla C(\S) = \relint ( \nabla C(\S))$ are technical but important, and we do not address these details in the present extended abstract although they will be considered in the full version. More relevant discussion can also be found in Abernethy et al. \cite{ACV11}.}.
\end{definition}\vspace{-4pt}

Fortunately, we need not provide a full description of the procedure of the APMM mechanism: The APMM is precisely a special case of a CLM! Indeed, the APMM framework can be described as a CLM $(\H,\O,\cost,\payout)$ where
\begin{equation} \label{eq:apmmdef}
	\H = \S (= \reals^n) \quad \quad
    \cost(\s, \s') = C(\s') - C(\s) \quad \quad % these were backwards!
    \payout(\s,\s';X) = \rhob(X) \cdot (\s' - \s).
\end{equation}
Hence we can think of APMM prediction markets in terms of our learning mechanism. Markets of this form are an important special class of CLMs -- in particular, we can guarantee that they are efficient to work with, as we show in the following proposition.
\begin{proposition}
An APMM $(\S, \O, \rhob, C)$ with a efficiently computable $C$ satisfies the EC and TT properties.
  \label{lem:tractable}
\end{proposition}\vspace{-4pt}
\begin{proof}
  Computing $\cost$ and $\payout$, defined in \eqref{eq:apmmdef}, requires simply evaluating $C(\cdot)$ at two inputs and evaluating $\rhob(\cdot)$ and taking a dot product, all of which are polynomial-in-$n$ operations. Hence an APMM satisfies the EC property. Furthermore, for a participant to make an optimal trade with belief $P$ under a budget constraint $B$, she must simply compute, for any fixed $\s'$,
  \[
    \mathop{\arg\max}_{\s \in \reals^n : C(\s) - C(\s') \leq B} \E_{X \sim P} [\rhob(X)] \cdot (\s - \s') - C(\s) + C(\s').
  \]
  The latter objective is a standard convex optimization problem in $n$ parameters which can be solved efficiently.
\end{proof}

We now ask, what is the learning problem that the participants of an APMM are trying to solve?  More precisely, when we think of an APMM as a CLM, does it implement a particular $L$?  

\begin{lemma} \label{lem:apmm}
  Let APMM $A = (\S,\O,\rhob,C)$ be given.  Then $A$ implements the GSR $L:\nabla C(\S)\times\O\to\reals$ defined by
  \begin{equation}
    L(\w;X) = D_{C^*}(\rhob(X),\w) + f(X),
    \label{eq:primal-apmm}
  \end{equation}
  where $C^*$ is the conjugate dual of the function $C$ and $f$ is arbitrary.
  \label{lem:primal-apmm}
  %NOTE: this does not actually need C^* = R; what we should really say is that it implements D_R for any R such that R* = C (I think)
\end{lemma}\vspace{-4pt}
\begin{proof}
  We analyze the profits for trades in $A$.  Let $\p_t = \nabla C(\s_t)$ be the instantaneous prices at time $t$.  The ex-post profit of the bet $(\s_t\mapsto\s_{t+1})$ is then
  \begin{align}
    \profit(\s_t,\s_{t+1};X)
    &= (\s_{t+1}-\s_t)\cdot\rhob(X) - C(\s_{t+1}) + C(\s_t)
    \nonumber
    \\
    &= (\s_{t+1}-\s_t)\cdot\rhob(X) - (\p_{t+1} \cdot \s_{t+1} - C^*(\p_{t+1})) + (\p_t \cdot \s_t - C^*(\p_t))
    \nonumber
    \\
    &= C^*(\p_{t+1}) + \s_{t+1}\cdot(\rhob(X)-\p_{t+1}) - C^*(\p_t) - \s_t\cdot(\rhob(X)-\p_t)
    \label{eq:profit-expression}
  \end{align}
  Now note that since $C$ is closed and convex, by duality we can write
  \begin{equation}\label{eq:duality-c}
    C(\s) = \sup_{\p\in\nabla C(\S)} \{\s\cdot\p - C^*(\p)\},
  \end{equation}
  for all $\s$.  Using standard techniques in conjugate duality theory, we can conclude that the $\sup$ is achieved for $\p = \nabla C(\s)$; we briefly sketch this argument here.  Since $C^*$ is the conjugate dual of $C$, we have
  \(
    C^*(\nabla C(\s)) = \sup_{\s' \in \reals^n} \s' \cdot \nabla C(\s) - C(\s').
  \)
  As this objective is unconstrained, we see that an optimal choice of $\s'$ is identically $\s$. This gives the following equality, 
  \begin{equation}\label{eq:duality-c*}
    C^*(\nabla C(\s)) =  \s \cdot \nabla C(\s) - C(\s).
  \end{equation}
  Of course, by reconciling equations~\eqref{eq:duality-c} and~\eqref{eq:duality-c*}, we see that one optimal choice of $\p$ is $\nabla C(\s)$; indeed this is the only choice, although we need not prove this statement.
  
  Continuing, since this $\p := \nabla C(\s)$ maximizes the objective in~\eqref{eq:duality-c}, we must have 
  \[
  0 = \v \cdot \nabla_\p \left(\s\cdot\p - C^*(\p)\right) = \v \cdot (\s - \nabla C^*(\p)),
  \]
  for any direction $\v \in \nabla C(\S)-\{\p\}$.  This is because $\p \in \nabla C(\S) \equiv \relint(\nabla C(\S))$ by assumption, so if the directional derivative were nonzero the objective would increase in the direction of $\v$ or $-\v$.  Now since $\rhob(\O)\subseteq\nabla C(\S)$ according to Definition~\ref{def:apmm}, we have $\rhob(X) - \p \in \nabla C(\S)-\{\p\}$ for all $\p$.  Thus,
  \begin{equation}\label{eq:duality-trick}
    \begin{aligned}
      \s_{t+1}\cdot(\rhob(X)-\p_{t+1}) &= \nabla
      C^*(\p_{t+1})\cdot(\rhob(X)-\p_{t+1}), \text{ and}
      \\
      \s_t\cdot(\rhob(X)-\p_t) &= \nabla
      C^*(\p_t)\cdot(\rhob(X)-\p_t).
    \end{aligned}
  \end{equation}
  Finally, applying \eqref{eq:duality-trick} and adding $C^*(\rhob(X)) - C^*(\rhob(X))$ to~\eqref{eq:profit-expression} yields
  \begin{equation}\label{eq:profit-divergence}
    \profit(\s_t,\s_{t+1};X)
    = D_{C^*}(\rhob(X),\p_t) - D_{C^*}(\rhob(X),\p_{t+1}),
  \end{equation}
  Note that if we had instead added $C^*(\pbar) - C^*(\pbar)$, where $\pbar = \E_{X\sim P}\,[\rhob(X)]$, we would see that the expected ex-post profit under $P$ is $D_{C^*}(\E[\rhob(X)],\p_t) - D_{C^*}(\E[\rhob(X)],\p_{t+1})$.  We now have
  \begin{equation*}
    \profit(\s_t,\s_{t+1};X) = L(\nabla C(\s_t); X) - L(\nabla C(\s_{t+1}); X),
  \end{equation*}
  since the $f(X)$ terms cancel; the implementation property thus follows with the surjective map $\varphi:\s\mapsto\nabla C(\s)$.
\end{proof}

There is another more subtle benefit to APMMs -- and, in fact, to most prediction market mechanisms in practice -- which is that participants make bets via purchasing of shares or share bundles. When a trader makes a bet, she purchases a contract bundle $\r$, is charged $C(\s + \r) - C(\s)$ (when the current quantity vector is $\s$), and shall receive payout $\rhob(X)\cdot\r$ if and when $X$ is realized. But at any point before $X$ is observed and trading is open, the trader can sell off this bundle, to the APMM or another trader, and hence neutralize her risk. In this sense bets made in an APMM are stateless, whereas for an arbitrary CLM this may not be the case: the wager defined by $(\w_t \mapsto \w_{t+1})$ can not necessarily be sold back to the mechanism, as the posted hypothesis may no longer remain at $\w_{t+1}$.

Given a learning problem defined by the GSR $L: \H \times \O \to \reals$, it is natural to ask whether we can design a CLM which implements this $L$ and has this ``share-based property'' of APMMs. More precisely, under what conditions is it possible to implement $L$ with an APMM?
% This leads us to the following key theoretical result.

\begin{theorem}
  For any divergence-based GSR $L(\w;X) = D_R(\rhob(X),\psi(\w)) + f(X)$, with $\psi:\H\to\H'$ one-to-one, $\H' = \relint(\H')$, and $\rhob(\O) \subseteq \psi(\H)$, there exists an APMM which implements $L$.
  % For every divergence-based GSR $L:\H\times\O\to\reals$ with $\psi:\H\to\H'$ one-to-one, there exists an APMM which implements $L$.
  \label{thm:primal-apmm}
\end{theorem}\vspace{-4pt}
\begin{proof}
  Recall the functions involved from Definition~\ref{def:div-based}: $\rhob: \O \to \H', f :\O \to \reals$, $\psi: \H\to\H'$, and finally $R: \H' \to \reals$ is a closed strictly convex function.  In order to construct an APMM we set $C = R^*$, the conjugate dual of $R$, and $\S = \dom(R^*)$.  Note that $\S = \reals^n$ by assumption.  Using standard results of conjugate duality we have $\nabla C(\S) = \dom (R) = \H'$, and since $R$ is closed and convex, we also have $C^* = (R^*)^* = R$.

  Since we have $\rhob(\O) \subseteq \H' = \nabla C(\S)$, we can construct an APMM $A = (\S, \O, \rhob, C)$.  By~\eqref{eq:profit-divergence} from Lemma~\ref{lem:primal-apmm}, we can write the profit of the bet $(\s_t\mapsto \s_{t+1})$ in $A$ as
  \begin{equation}
    \profit(\s_t,\s_{t+1};X)
    = D_R(\rhob(X),\nabla C(\s_t)) - D_R(\rhob(X),\nabla C(\s_{t+1})).
  \end{equation}
  As $\psi$ is one-to-one by assumption, we can now write
  \begin{equation*}
    \profit(\s_t,\s_{t+1};X) = L(\psi^{-1}(\nabla C(s_t));X) - L(\psi^{-1}(\nabla C(s_{t+1}));X),
  \end{equation*}
  since the $f(X)$ terms cancel.  Finally, since $\nabla C: \S\to\H'$ and $\psi^{-1}:\H'\to\H$ are surjective, we see that $A$ implements $L$ with map $\varphi \equiv \psi^{-1} \circ \nabla C$.
\end{proof}

We point out that if an APMM implements some arbitrary $L$, not necessarily the canonical $L_0$ established in Lemma~\ref{lem:apmm},  then $L$ is effectively equivalent to $L_0$ and, hence, is divergence based. This fully specifies the class of problems solvable using APMMs.

\begin{theorem}
  If APMM $(\S,\O,\rhob,C)$ implements a GSR $L:\H\times\O\to\reals$, then $L$ is divergence-based.
  \label{thm:char}
\end{theorem}\vspace{-8pt}
\begin{proof}
  Given that $(\S, \O, \rhob, C)$ implements $L$, we know there exists some surjective map $\varphi:\S\to\H$ so that $\profit(\s,\s';X) = L(\varphi(\s);X) - L(\varphi(\s');X)$ for all $\s,\s'\in\S$. Of course, applying Lemma~\ref{lem:apmm} we now have,
  \begin{equation*}
    D_R(\rhob(X),\nabla C(\s)) - D_R(\rhob(X),\nabla C(\s')) = \profit(\s,\s';X) = L(\varphi(\s);X) - L(\varphi(\s');X),
  \end{equation*}
  where $R=C^*$.
  Focusing on $L(\cdot;\cdot)$ as a function of $\s$, and fixing $\s'$ arbitrarily, we see that
  \begin{align*}
    L(\varphi(\s);X)
    & \equiv D_R(\rhob(X),\nabla C(\s)) + \bigl(L(\varphi(\s');X) - D_R(\rhob(X),\nabla C(\s'))\bigr) \\
    & \equiv D_R(\rhob(X),\nabla C(\s)) + f(X)
  \end{align*}
  for some $f:\O\to\mathbb{R}$, since $L(\varphi(\s);X)$ cannot depend on the unbound variable $\s'$.  Furthermore, for any $\w,\s$ such that $\w = \varphi(\s)$, we must have
  \begin{equation*}
    L(\w;X) = D_R(\rhob(X),\nabla C(\s)) + f(X).
  \end{equation*}
Now since $\varphi$ is surjective, there exists a map $\tilde\varphi$ with $\varphi \circ \tilde\varphi = \text{id}_{\H}$.  Hence,
\begin{equation*}
  L(\w;X) \equiv D_R(\rhob(X),\psi(\w)) + f(X),
\end{equation*}
where $\psi = \nabla C \circ \tilde \varphi$.
\end{proof}\vspace{-4pt}

Theorem~\ref{thm:char} establishes a strong connection between
prediction markets and a natural class of GSRs.  One interpretation of
this result is that any GSR based on a Bregman divergence has a ``dual''
characterization as a share-based market, where participants buy and sell shares
rather than directly altering the share prices (the hypothesis).  This
has many advantages for prediction markets, not least of which is that
shares are often easier to think about than the underlying hypothesis
space.

Our notion of a CLM offers another interpretation.  In light of
Proposition~\ref{lem:tractable}, any machine learning problem whose
hypotheses can be evaluated in terms of a divergence leads to a
tractable crowdsourcing mechanism, as was the case in Section~\ref{sec:compression}.
Moreover, this theorem does not preclude
efficient yet non-divergence-based loss functions as we see in the next section.
%   For example, we see now that the
% GSR from Section~\ref{sec:compression} is implementable by an APMM.
% As we will see in Section~\ref{sec:y-mkt}, Theorem~\ref{thm:char} is
% also very relevant when the goal is simply to predict labels (as in
% the Netflix prize).  Moreover, this theorem does not preclude
% efficient yet non-divergence-based machine learning problems as we see in the next section.

% \begin{corollary}[]
%   The loss function in section~\ref{sec:compression} is implemented by
%   the Logarithmic Market Scoring Rule APMM $(\Delta_n,\Delta_n,\rhob:i\mapsto\mathbf{e}_i,(-H)^*)$.
%   \label{cor:compression}
% \end{corollary}

\section{Example CLMs for Typical Machine Learning Tasks} \label{sec:examples}
\newcommand{\al}{{\boldsymbol\alpha}}
\newcommand{\test}{\textsf{test}}
\newcommand{\train}{\textsf{train}}
% 
% A possible general ML framework:
% 
% \begin{itemize}
% \item Given training data $X_0 = \{\x_i,y_i\}_{i\in\train}$
% \item Outcome is the test data $X = \{\x_i,y_i\}_{i\in\test}$
% \item Loss $L(\w;X)$ is convex in $\w$, plus whatever other properties
%   we need for Theorem~\ref{thm:canon} to hold.
% \end{itemize}
% 
% \subsection{SVM}
% \begin{itemize}
% \item $W = \{\al\} = \reals^{|\textsf{train}|}$, the dual variables (weighs on the support vectors)
% \item $X = \{\x_i,y_i\}_{i\in\textsf{test}}$, the test data
% \item $L(\al;X) = C \sum_{j\in\train} (\alpha_j y_j)^2 K(\x_j,\x_j) + \sum_{i\in\textsf{test}} \left[1 - \sum_{j\in\textsf{train}} \alpha_j y_i y_j K(\x_j,\x_i)\right]_+$
% \end{itemize}
% 
\newcommand{\X}{\mathcal{X}}
\newcommand{\Y}{\mathcal{Y}}

\paragraph{Regression.} We now construct a CLM for a typical regression problem. We let $\H$
be the $\ell_2$-norm ball of radius $1$ in $\reals^d$, and we shall
let an outcome be a batch of a data, that is $X := \{(\x_1,y_1),
\ldots, (\x_n,y_n)\}$ where for each $i$ we have $\x_i \in \reals^d$,
$y_i \in [-1,1]$, and we assume $\|\x_i\|_2 \leq 1$. We construct a
GSR according to the mean squared error, $L(\w;
\{(\x_i,y_i)\}_{i=1}^n) = \frac \alpha {2n} \sum_{i=1}^n (\w\cdot\x_i
- y_i)^2$ for some parameter $\alpha > 0$. It is worth noting that $L$ is not divergence-based.

In order to satisfy the escrow property (ES), we can set
$\cost(\w,\w') := 2\alpha\|\w - \w'\|_2$ because the function
$L(\w;X)$ is $2\alpha$-lipschitz with respect to $\w$ for any $X$. To
ensure that the CLM is $L$-incentivized, we must set
$\payout(\w,\w';X) := \cost(\w,\w') + L(\w; X) - L(\w';X)$.

If we set the initial hypothesis $\w_0 = \0$, it is easy to check
that $\worstcaseloss = \alpha/2$. It remains to check whether this CLM
is tractable. It's clear that we can efficiently compute $\cost$ and
$\payout$, hence the EC property holds. Given how $\cost$ is defined,
it is clear that the set $\{ \w' : \cost(\w,\w') \leq B \}$ is just an
$\ell_2$-norm ball. Also, since $L$ is convex in $\w$ for each $X$, so
is the function $\E_{X\sim P} \bigl[\profit(\w,\w',X)\bigr]$ for every $P$. A budget-constrained profit-maximizing participant
must simply solve a convex optimization problem, and hence the TT
property holds.
% \begin{itemize}
% \item want to predict $\y \in \Y^n \subset \mathbb{R}^n$ from
%   $\{\x_i\}_{i\in[n]} \subset \X \subset \mathbb{R}^d$
% \item wish to minimize $\sum (\hat y_i - y_i)^2$
% \item want a linear predictor $\hat y_i = \w \cdot \x_i$
% \item have some bounds on the norms of $y_i, \x_i$ (?)
% \end{itemize}
% 
% Given the $\{\x_i\}$, let $X = [\x_1 \cdots \x_n]^\top$; then we can write our loss
% \begin{equation*}
%   L(\w; X,\y) = \frac 1 n \sum_{i=1}^n (\w\cdot\x_i - y_i)^2
%   = \|X\w\|^2 - 2(X\w)\cdot\y + \|\y\|^2.
% \end{equation*}
% 
% \begin{align*}
%   \cost(\w,\w')
%   &= \max_{X\in\X^n,\y\in\Y^n} \left(L(\w'; X,\y) - L(\w; X,\y)\right)
%   \\
%   &= n \max_{\x\in\X,y\in\Y} \left((\w'\cdot\x - y)^2 - (\w\cdot\x - y)^2\right)
%   \\
%   &= n \max_{\x\in\X,y\in\Y} \left((\w'\cdot\x)^2 - (\w\cdot\x)^2 - 2y\x\cdot(\w'-\w)\right)
% \end{align*}
% 
% \begin{align*}
%   \w_{t+1} 
%   &= \argmin_{\w\in\H} \Bigl\{ \E_{(X,\y)\sim P}
%   \bigl[ \|X\w_t\|^2 -\|X\w\|^2 - 2(X\w_t-X\w)\cdot\y \bigr] :\: \cost(\w_t,\w)\leq B \Bigr\}.
%   \\
%   &= \argmin_{\w\in\H} \Bigl\{ \E_{(X,\y)\sim P}
%   \bigl[ 2(X\w)\cdot\y - \|X\w\|^2 \bigr] :\: \cost(\w_t,\w)\leq B \Bigr\}.
% \end{align*}

\paragraph{Betting Directly on the Labels.} Let us return our attention to the Netflix Prize model as discussed in
the Introduction. For this style of competition a host releases a dataset for a given prediction task. The host then
requests participants to provide predictions on a specified set of
instances on which it has correct labels. For every submission the
agent computes an error measure, say the MSE, and reports this to the
participants. Of course, the correct labels are withheld throughout.

Our CLM framework is general enough to apply to this problem framework
as well. Define $\H = \O = K^m$ where $K\subseteq\reals$ bounded is
the set of valid labels, and $m$ is the number of requested test set
predictions. For some $\w \in \H$ and $\y \in \O$, $\w(k)$ specifies
the $k$th predicted label, and $\y(k)$ specifies the true label. A
natural scoring function is the total squared loss, $L(\w; \y) :=
\sum_{k=1}^m (\w(k) - \y(k))^2$. Of course, this approach is quite
different from the Netflix Prize model, in two key respects: (a) the
participants have to wager on their predictions and (b) by
participating in the mechanism they are required to \emph{reveal their
  modification} to all of the other players. Hence while we have
structured a competitive process the participants are de facto forced
to collaborate on the solution.

A reasonable critique of this collaborative mechanism approach to a
Netflix-style competition is that it does not provide the instant
feedback of the ``leaderboard'' where individuals observe performance improvements in real time. However, we can
adjust our mechanism to be online with a very simple modification of
the CLM protocol, which we sketch here. Rather than make payouts in a
large batch at the end, the competition designer could perform a
mini-payout at the end of each of a sequence of time intervals. At
each interval, the designer could select a (potentially random) subset
$S$ of user/movie pairs in the remaining test set, freeze updates on
the predictions $\w(k)$ for all $k \in S$, and perform payouts to the
participants on only these labels. What makes this possible, of
course, is that the generalized scoring rule we chose decomposes as a
sum over the individual labels.

With this online approach just discussed, let us end with one final observation. Given that some firm such as Netflix would like to make good predictions on a given data source, the firm could potentially rely \emph{entirely} on the advice from a CLM. The firm could post batches of data on which it does not have labels and ask the participants to provide predictions via the CLM. The firm could then pay out according to a small sample which are manually labeled or whose labels are received in the mean time. But by not revealing on which subset the labels will arrive, the firm receives potentially good predictions on the full set. This could provide a valuable market between small firms which have machine learning needs and ``freelance'' machine learning bounty hunters. 

\paragraph{Acknowledgments.} We gratefully acknowledge the support of the NSF under award DMS-0830410, a Google University Research Award, and the National Defense Science and Engineering Graduate (NDSEG) Fellowship, 32 CFR 168a.

% 
% \subsubsection*{References}
% 
% References follow the acknowledgments. Use unnumbered third level heading for
% the references. Any choice of citation style is acceptable as long as you are
% consistent. It is permissible to reduce the font size to `small' (9-point) 
% when listing the references. {\bf Remember that this year you can use
% a ninth page as long as it contains \emph{only} cited references.}

\bibliographystyle{plain}
\bibliography{nips-crowd}

\end{document}